%% file: arxiv.tex
\documentclass{article}
\usepackage[preprint]{neurips_2025}

\usepackage[utf8]{inputenc} 
\usepackage[T1]{fontenc}    
\usepackage{hyperref}       
\usepackage{url}            
\usepackage{booktabs}       
\usepackage{amsfonts}       
\usepackage{nicefrac}       
\usepackage{microtype}      
\usepackage{xcolor}         

\usepackage{multirow}
\usepackage{makecell}
\usepackage{tcolorbox}
\usepackage{float}
\usepackage{pifont}
\usepackage{colortbl}
\usepackage{graphicx}
\usepackage{subcaption}
\usepackage{amssymb,amsmath,amsthm}
\usepackage{paralist}

\usepackage{mathtools}

\usepackage{cleveref}
\usepackage{enumitem}
\usepackage{arydshln}
\usepackage{listings}

\newcommand{\ourbench}{OverLayBench}
\newcommand{\ourbaseline}{CreatiLayout-AM}
\newcommand{\ourmetric}{OverLayScore}

\title{\ourbench{}: A Benchmark for Layout-to-Image Generation with Dense Overlaps}

\author{%
 Bingnan Li$^{1*}$~~~~Chen-Yu Wang$^{1*}$~~~~Haiyang Xu$^{1*}$~~~~Xiang Zhang$^1$~~~~Ethan Armand$^1$ \\
 \textbf{Divyansh Srivastava$^1$~~~~Xiaojun Shan$^1$~~~~Zeyuan Chen$^1$~~~~Jianwen Xie$^2$~~~~Zhuowen Tu$^1$} \\
 $^1$UC San Diego~~~~$^2$Lambda,~Inc \\
 {\small $^*$equal contribution}
}

\begin{document}

\maketitle

\vspace{-12pt}
\begin{figure}[h]
  \centering
  \includegraphics[width=1.0\linewidth]{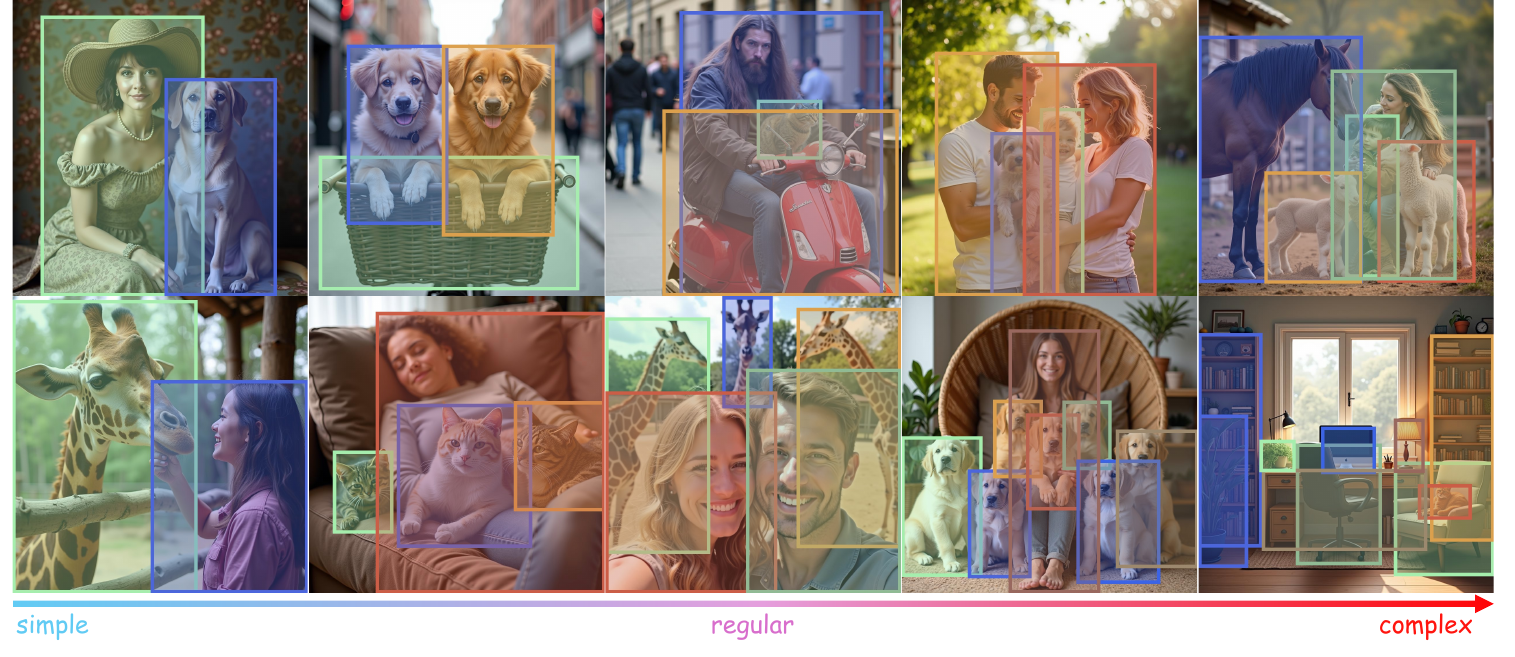}
  \caption{Examples from \ourbench{} with difficulty increasing from left to right.}
  \label{fig:teaser}
\end{figure}

\begin{abstract}
    Despite steady progress in layout-to-image generation, current methods still struggle with layouts containing significant overlap between bounding boxes. We identify two primary challenges: (1) large overlapping regions and (2) overlapping instances with minimal semantic distinction. Through both qualitative examples and quantitative analysis, we demonstrate how these factors degrade generation quality. To systematically assess this issue, we introduce \ourmetric{}, a novel metric that quantifies the complexity of overlapping bounding boxes. Our analysis reveals that existing benchmarks are biased toward simpler cases with low \ourmetric{} values, limiting their effectiveness in evaluating model performance under more challenging conditions. To bridge this gap, we present \ourbench{}, a new benchmark featuring high-quality annotations and a balanced distribution across different levels of \ourmetric{}. As an initial step toward improving performance on complex overlaps, we also propose \ourbaseline{}, a model fine-tuned on a curated amodal mask dataset. Together, our contributions lay the groundwork for more robust layout-to-image generation under realistic and challenging scenarios. \textit{Project link: \href{https://mlpc-ucsd.github.io/OverLayBench}{https://mlpc-ucsd.github.io/OverLayBench}}.
\end{abstract}

\section{Introduction}

\begin{figure}[ht]
  \centering
  \includegraphics[width=0.98\linewidth]{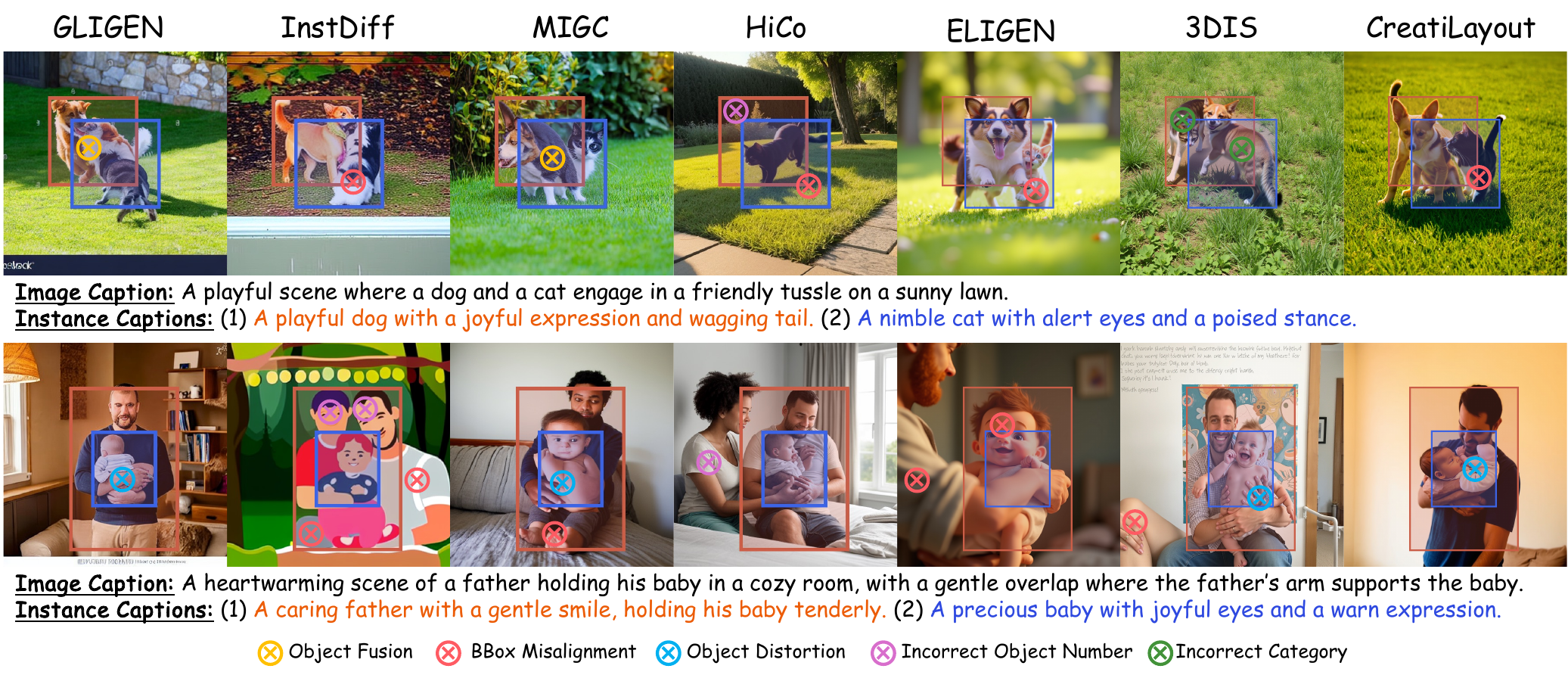}
  \caption{Failure cases from state-of-the-art layout-to-image models. Each row presents an example with overlapping instances, and image captions are shown below. More examples and more detailed failure descriptions can be found in~\Cref{supp:failure_case}.}
  \vspace{-12pt}
  \label{fig:failure-cases}
\end{figure}

With the advancement of text-to-image generative models~\citep{ramesh2021zero,nichol2021glide,rombach2022high,chen2023pixart,xue2024raphael}, there has been growing interest in \textit{controllable image generation}~\citep{li2023gligenopensetgroundedtexttoimage,zhang2023adding}. A recent line of work proposes generating images conditioned on layouts, commonly referred to as \textbf{Layout-to-Image (L2I)} generation, which allows users to directly specify spatial locations~\citep{xie2023boxdiff,wang2024instancediffusion,li2023gligenopensetgroundedtexttoimage} and object counts~\citep{binyamin2024make,yang2023reco} in the generated outputs.  While existing frameworks ~\citep{xie2023boxdiff,wang2024instancediffusion,li2023gligenopensetgroundedtexttoimage} can achieve satisfactory spatial and numerical control over image generation, these approaches fail to generate distinct, coherent objects when multiple bounding boxes overlap in layout and their associated categories are semantically similar. As illustrated in~\Cref{fig:failure-cases}, such scenarios lead to artifacts including object blending, spatial ambiguity, and visual distortion.

To quantify the effect of bounding box overlap on generation, we introduce \textbf{\ourmetric{}} (\Cref{eq:overlayscore}), a simple metric that captures the difficulty of generation based on spatial and semantic overlap between bounding boxes. \ourmetric{} is computed as the sum of IoUs for all instance pairs, weighted by their semantic similarity, measured via the dot product of CLIP embeddings of instance annotations. We empirically demonstrate that generation quality degrades with higher \ourmetric{}, i.e., large overlap between bounding boxes and high category-level semantic similarity. Henceforth, we will interchangeably refer to layouts with high \ourmetric{} score as complex layouts and low \ourmetric{} score as simple layouts.

Existing benchmarks for Layout-to-Image (L2I) generation~\citep{hico, creatilayout} primarily focus on image quality, offering limited evaluation of complex layouts or the accuracy of generated instance relationships. Our analysis reveals a strong bias towards simple layouts in these benchmarks, which restricts their utility in assessing model performance under more challenging, realistic scenarios. To address this gap, we introduce \textbf{\ourbench{}}, a new benchmark specifically designed to evaluate L2I models on their ability to reconstruct complex layouts and instance-level relationships. \ourbench{} features rich annotations such as detailed image and dense instance captions, enrichment of complex images with higher \ourmetric{}, improved semantic grounding of bounding boxes using Qwen~\citep{qwen}, and quality instance relationships for evaluation. We conduct extensive evaluations of state-of-the-art L2I models~\citep{gligen, instancediffusion, migc, hico, 3dis, eligen, creatilayout} on \ourbench{} and verify their effectiveness in addressing layout-level challenges. These results provide strong baselines and highlight areas for improvement.

Finally, we demonstrate that fine-tuning CreatiLayout~\citep{creatilayout} with amodal mask supervision on complex layouts helps mitigate generation artifacts caused by instance occlusion. This new baseline, \textbf{\ourbaseline{}}, provides initial evidence that explicit mask-level guidance improves generation quality under high-overlap conditions, offering a promising direction for future research.

Our contributions are summarized as follows:
(1) We propose \textbf{\ourmetric{}}, a novel metric that empirically quantifies the difficulty of L2I generation by measuring the IoU and semantic similarity between bounding boxes in layouts;
(2) We introduce \textbf{\ourbench{}}, a challenging benchmark with high-quality annotations and balanced difficulty distribution, designed to evaluate complex relationships between instances in layouts;
(3) We demonstrate that training with amodal masks helps alleviate generation artifacts in overlapping regions. Specifically, we present a simple yet effective baseline that aligns attention maps in diffusion models with amodal mask supervision.

\section{Related Work}
\subsection{Layout-to-Image Generation}
Generative models have recently become highly popular~\citep{omnicontrolnet,depr,yolocount,layyourscene,bdm,dolfin,xdancer}, and controllable generation is attracting growing interest. In particular, Layout-to-Image (L2I) generation has gained attention as it enables structured and spatially grounded image synthesis.

Prior work on L2I generation has primarily focused on fine-tuning foundational text-to-image (T2I) generative models~\citep{flux2024, sd3, podell2023sdxl}, introducing various techniques to inject layout conditioning into pre-trained architectures. U-Net-based approaches~\citep{gligen, migc, instancediffusion, ifadapter, hico, zheng2023layoutdiffusion} typically incorporate layout information through layer insertions (either in series or parallel) or by manipulating attention maps and masks. While these methods have shown promising results, their generation quality is often constrained by the representational capacity of U-Net backbones. More recent methods~\citep{eligen, creatilayout} leverage powerful diffusion Transformer (DiT) architectures~\citep{peebles2023dit}, fine-tuning models such as Flux~\citep{flux2024} or SD3~\citep{sd3}, and achieve improved image fidelity. These layouts can be provided by the user or generated automatically from text using text-to-layout models~\citep{feng2023layoutgpt,llmblueprint,lian2023llm,srivastava2025lay}, forming a two-stage pipeline for controllable image synthesis. In parallel, a growing line of training-free approaches~\citep{boxdiff, llmblueprint, rpg, groundit, trainigfreeregional, rag} has emerged, which utilize guidance mechanisms within diffusion models to enforce spatial constraints without additional training. However, these models often treat layout information as a soft constraint, which limits their ability to strictly adhere to spatial specifications.

Despite these advancements, existing methods struggle to generate coherent and distinguishable objects when multiple bounding boxes overlap, particularly when the associated categories are semantically similar. Such scenarios frequently lead to confusion, ambiguity, visual distortion, or artifacts in the generated images. To address the gap, we study Layout-to-Image (L2I) generation under bounding box overlap in layouts and highlight it as a novel and underexplored challenge that demands dedicated investigation.

\subsection{Layout-to-Image Benchmarks}

The most widely used benchmark for L2I generation is COCO~\citep{coco}, which provides image-text pairs annotated with entity bounding boxes (bboxes) and simple category labels. Although COCO and its variants are commonly used for evaluating L2I models, they lack detailed image and instance-level captions, which limits their utility for training and evaluating in more semantically rich generation tasks~\citep{feng2023ranni}. To address this shortcoming, many works manually augment COCO with additional instance-level captions to support more comprehensive evaluation.

Recent efforts~\citep{hico, creatilayout} have introduced new layout benchmarks to offer more robust and holistic assessments. HiCo~\citep{hico} proposes HiCo-7k, a benchmark containing 7,000 carefully curated samples from GriT-20M~\citep{peng2023kosmos}, while CreatiLayout~\citep{creatilayout} introduces LayoutSAM, comprising 5,000 samples selected from the SAM-1B~\citep{kirillov2023segment} dataset. Both benchmarks employ image/entity filtering and generate detailed annotations. However, they rely on GroundingDINO~\citep{liu2023groundingdino} for bounding box extraction, which, according to its authors, exhibits limited recognition performance and frequently produces false positives in detection outputs (as discussed in \Cref{sec:data_curation}). HiCo-7k partially addresses these issues through manual human curation, whereas LayoutSAM is generated without human intervention.

In contrast, \ourbench{} leverages Qwen2.5-VL-32B~\citep{bai2025qwen25vl}, which surpasses GroundingDINO in benchmark performance and our own qualitative evaluation. Additionally, we incorporate human curation to further ensure data quality. Most importantly, unlike previous benchmarks that primarily feature simple or regular layouts, our benchmark includes significantly more complex and challenging layouts, providing a more balanced distribution of layout difficulty and enabling more rigorous evaluation of L2I models.

\section{\ourmetric{}}

\begin{figure}[t]
  \centering
   \begin{minipage}{\textwidth}
        \centering
        \begin{subfigure}{\linewidth}
            \centering
            \includegraphics[width=\linewidth,trim=0 1em 0 0,clip]{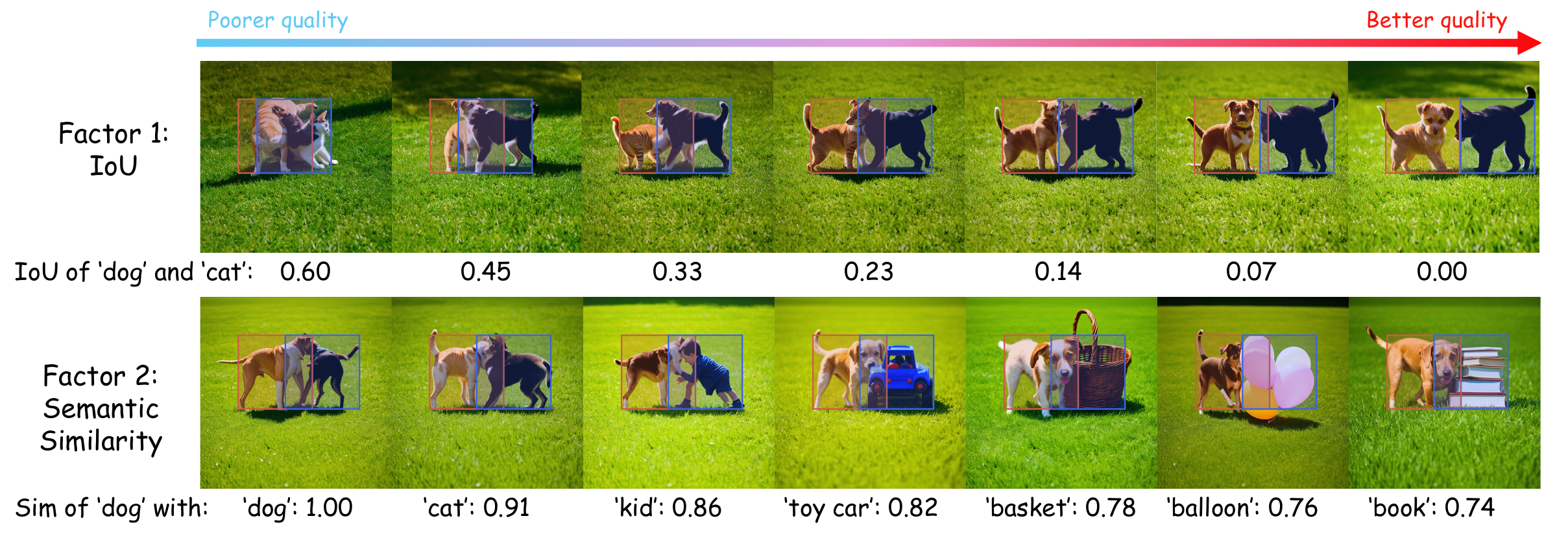}
            \caption{}\label{fig:motivation_a}
        \end{subfigure}
    \end{minipage}
    \begin{minipage}{\textwidth}
        \centering
        \vspace{-.5em}
        \begin{subfigure}{0.49\linewidth}
            \centering
            \includegraphics[width=\linewidth,trim=0 1em 0 0,clip]{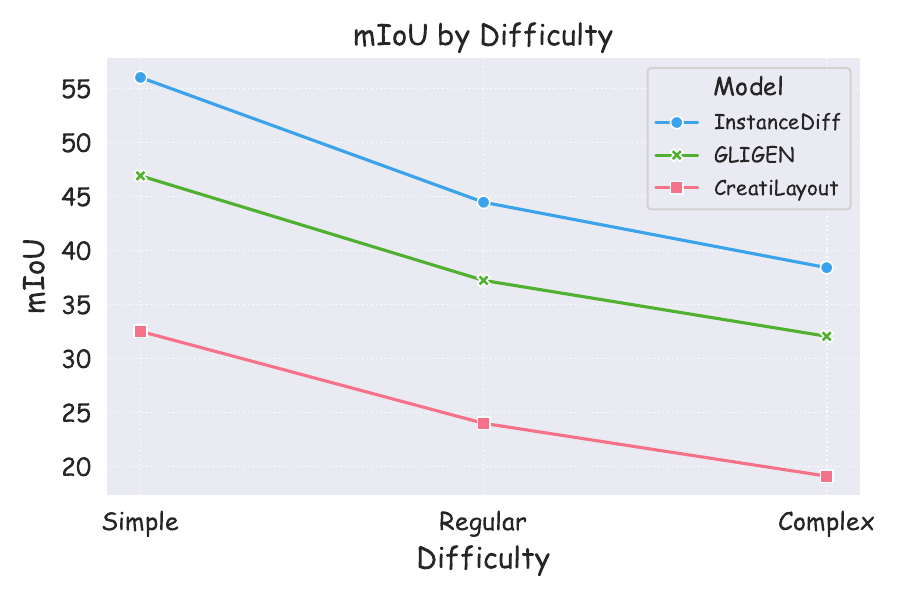}
            \caption{}\label{fig:motivation_b}
        \end{subfigure}
        \hfill
        \begin{subfigure}{0.49\linewidth}
            \centering
            \includegraphics[width=\linewidth,trim=0 1em 0 0,clip]{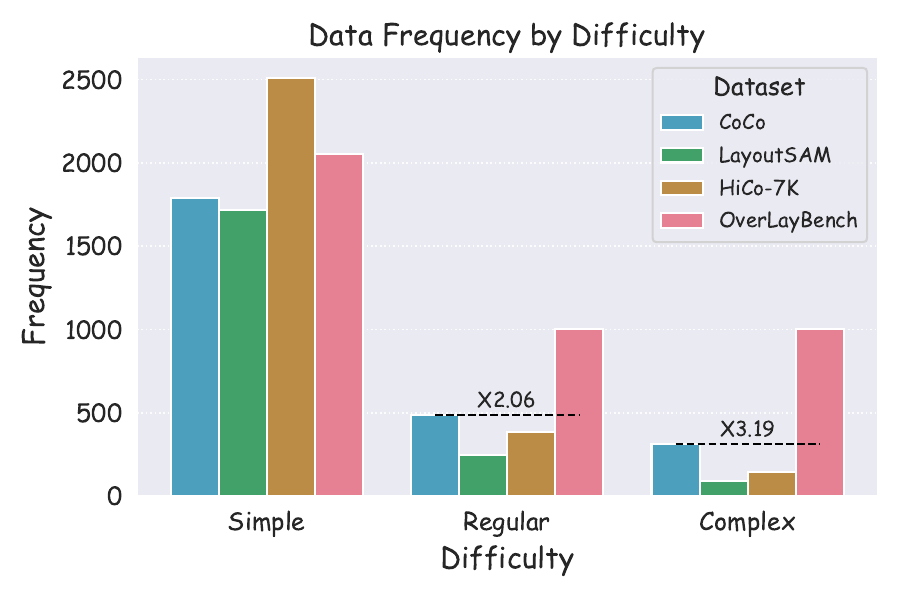}
            \caption{}\label{fig:motivation_c}
        \end{subfigure}
    \end{minipage}
  \caption{(a) Image quality of CreatiLayout across varying levels of bounding box IoU and semantic similarity, from poorer to better; (b) Performance comparison of three L2I models on toy COCO samples, grouped by layout difficulty using \ourmetric{}—higher difficulty consistently leads to lower mIoU; (c) Distribution of layout difficulty across COCO, LayoutSAM, HiCo-7k, and our proposed benchmark, \ourbench{} (introduced in detail in \Cref{sec:data_curation}).}
  \label{fig:motivation}
\end{figure}

We evaluate the impact of overlap on L2I generation performance from two perspectives: spatial and semantic. \Cref{fig:motivation} (a) presents the L2I generation results in a simplified two object setting, from which we derive two key observations: (1) as the Intersection-over-Union (IoU) between bounding boxes increases, the output quality of state-of-the-art L2I models deteriorates; and (2) given the same IoU, a higher semantic similarity between the instance captions of overlapping boxes further degrades the generation quality. These findings indicate that both spatial and semantic overlaps introduce complexities that negatively affect L2I generation.

Building on our observations, we propose \ourmetric{}, a metric designed to quantify the difficulty of L2I generation arising from overlapping elements within a layout. Formally, given a layout with $K$ objects, let $p_k$ and $B_k$ denote the instance caption and normalized bounding box for the $k$-th object, respectively. We define the metric as:
\begin{equation} \label{eq:overlayscore}
    \mathtt{\ourmetric{}} = \sum_{(i, j):\;\text{IoU}(B_i, B_j) > 0} \text{IoU}(B_i, B_j) \cdot \cos \big(\langle p_i, p_j \rangle \big),
\end{equation}
where $\cos \big(\langle p_i, p_j \rangle \big)$ is the CLIP-based cosine similarity between instance caption $p_i$ and $p_j$. The summation captures the spatial-semantic entanglement of all overlapping object pairs in the layout. A higher \ourmetric{} indicates greater expected difficulty for an L2I model to faithfully generate an image that conforms to both the layout and the associated instance-level semantics.

To validate the effectiveness of \ourmetric{}, we evaluate common L2I models on a subset of the COCO dataset~\citep{coco}. Layouts are extracted using the dataset's bounding box and category annotations, and filtered to include scenes with 2 to 10 objects. Based on their \ourmetric{}, these layouts are categorized into three difficulty levels -- simple, regular, and complex. From each category, we randomly sample 100 layouts for evaluation. We then assess the performance of three representative L2I models, i.e., GLIGEN~\citep{gligen}, InstanceDiffusion~\citep{instancediffusion}, and CreatiLayout~\citep{creatilayout}. As illustrated in \Cref{fig:motivation} (b), the performance of all models consistently declines as \ourmetric{} increases, demonstrating that \ourmetric{} effectively reflects the difficulty of generating images from overlapping layouts.

We further apply \ourmetric{} to several widely used L2I benchmarks--COCO~\citep{coco}, HiCo~\citep{hico}, and LayoutSAM~\citep{creatilayout}, and visualize the score distribution in \Cref{fig:motivation} (c). We observe that the majority of samples fall within the low-difficulty regime, indicating a strong imbalance in existing benchmarks. This skew limits their ability to evaluate model performance in more complex overlapping scenarios.

\section{\ourbench{}}
The analysis in the previous section highlights a key limitation of existing L2I benchmarks--they are heavily skewed toward low-\ourmetric{} samples, which restricts their ability to evaluate model performance under challenging layout conditions. To overcome this limitation, we introduce \ourbench{}—a new benchmark specifically curated to assess L2I models on complex and overlapping layouts. By carefully selecting images across a broad range of \ourmetric{} values, \ourbench{} provides a more balanced and comprehensive evaluation set, enabling rigorous evaluation of model robustness in spatially and semantically complex layouts.

\subsection{Dataset Curation}
\label{sec:data_curation}

\begin{figure}[t]
  \centering
  \includegraphics[width=\linewidth]{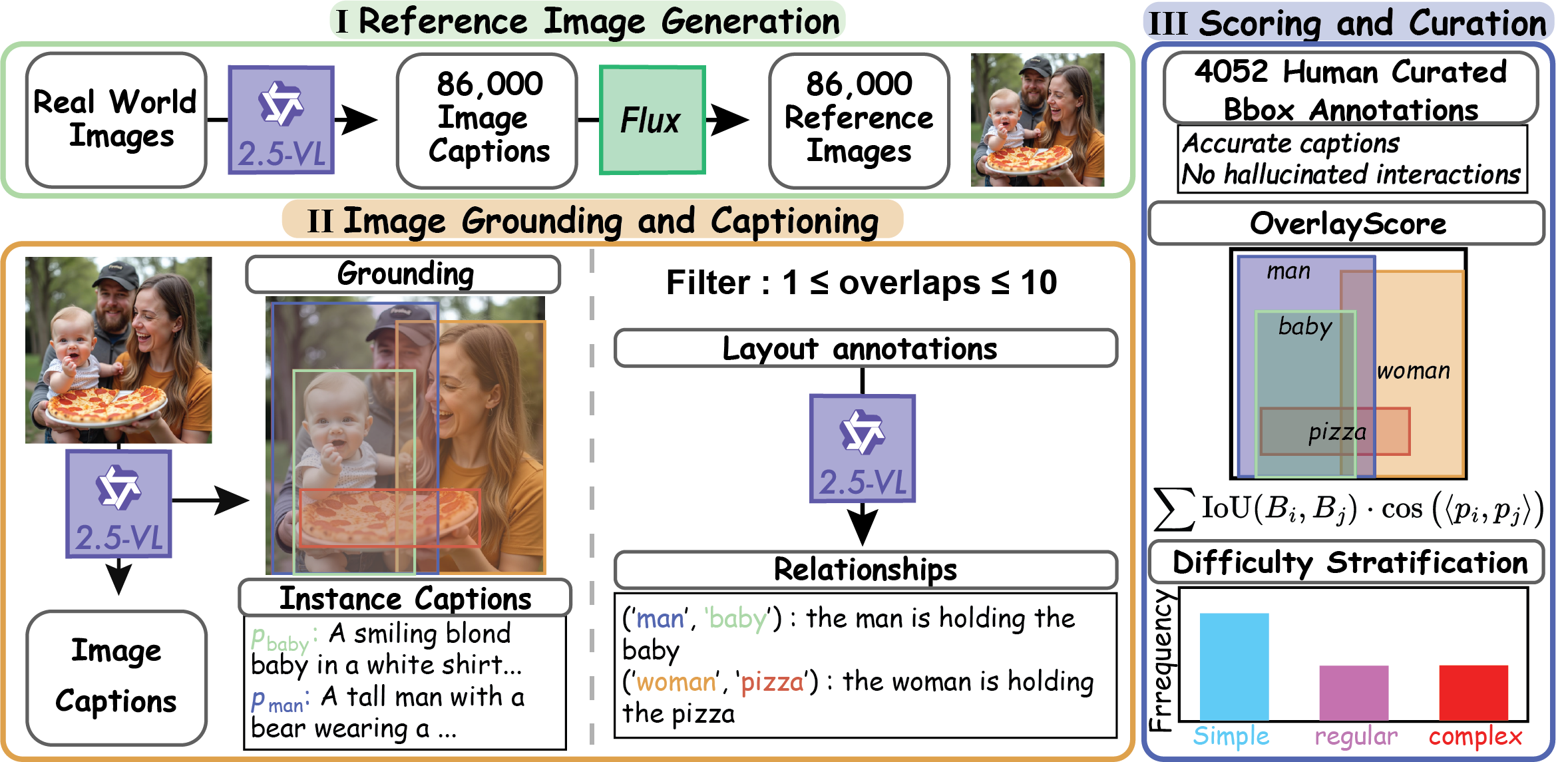}
  \caption{An overview of the data curation pipeline for \ourbench{}.}
  \label{fig:dataset-curation}
\end{figure}

An overview of our data curation pipeline is presented in \Cref{fig:dataset-curation}, comprising three key stages. In Stage I, we use Flux to generate reference images based on captions extracted from real-world images. Stage II leverages a vision-language model (VLM) to extract both image and instance-level descriptions, along with inter-instance relationships. Finally, Stage III involves a human curation process to filter out unrealistic generation and balance the distribution across difficulty levels.

\subsubsection{Reference Image Generation}
We begin by extracting image captions from the COCO~\citep{coco} training set using Qwen2.5-VL-7B~\citep{bai2025qwen25vl}. These captions are then used to generate a diverse set of image candidates with Flux.1-dev~\citep{flux2024}. By leveraging captions derived from real-world images, we ensure that the generated content and corresponding layouts are both natural and realistic. In total, we collect approximately 86,000 generated images paired with their corresponding captions.

\subsubsection{Image Grounding and Captioning}

\paragraph{Step 1: Image Caption Refinement} Although Flux demonstrates strong image generation capabilities, the generated images do not always perfectly align with the input captions. To improve semantic consistency, we perform an additional captioning pass on all generated images using Qwen‑2.5‑VL‑7B to produce refined global image captions.

\paragraph{Step 2: Instance Grounding} Qwen~\citep{qwen} has demonstrated superior grounding performance compared to models commonly used in existing L2I benchmarks, such as GroundingDINO~\citep{liu2023groundingdino}. We leverage Qwen to detect and describe all foreground objects in each image. Based on the grounding results, we retain only images that contain one to ten valid overlapping bounding box pairs. A bounding box pair is considered valid if it satisfies both: (1) an IoU greater than 5\%, and (2) an intersection area exceeding 1\% of the total image area. After this step, each image contains a global description and local descriptions for all detected instances.

\paragraph{Step 3: Relationship Extraction} In addition to image-level and instance-level captions and bounding boxes, we further prompt Qwen to generate pairwise relationship phrases between overlapping instances. These phrases capture  both spatial and semantic relationships, providing a richer annotation signal crucial for evaluating model performance on inter-instance relationship in complex layouts.

\subsubsection{Scoring and Curation}

To ensure high-quality and reliable annotations while minimizing hallucinations from VLMs, we perform thorough manual verification and discard all invalid cases. Specifically, we assess the accuracy of each bounding box, the alignment between image content and both global and instance-level captions, and the validity of relationship descriptions based on available textual inputs. This rigorous process ensures that \ourbench{} remains free from hallucinations. After validation, we compute the \ourmetric{} score for each example and retain a curated dataset of 2,052 simple, 1,000 regular, and 1,000 complex layouts.

\subsection{Benchmark Metrics}

We introduce two novel metrics tailored to overlapping scenarios: \textbf{O-mIoU} (Overlap-mIoU), which computes the mIoU within the ground-truth overlap regions and the corresponding predicted regions. By isolating shared areas between instances, O-mIoU provides a more sensitive and discriminative measure of fidelity in occluded or entangled regions than standard global mIoU; $\text{\bf SR}_\text{\bf R}$ (Success Rate of Relationship), which reports the percentage of object pairs whose predicted spatial relationships match the ground truth. It offers an interpretable, relationship-level measure of success.

In addition to these two metrics, we adopt commonly used evaluation measures from prior work~\citep{eligen, creatilayout}, including mIoU, CLIP~\citep{radford2021clip}, $\text{SR}_\text{E}$ (Success Rate of Entity), and FID~\citep{heusel2017fid}.

\begin{figure}[thb]
    \centering
    \includegraphics[width=\textwidth]{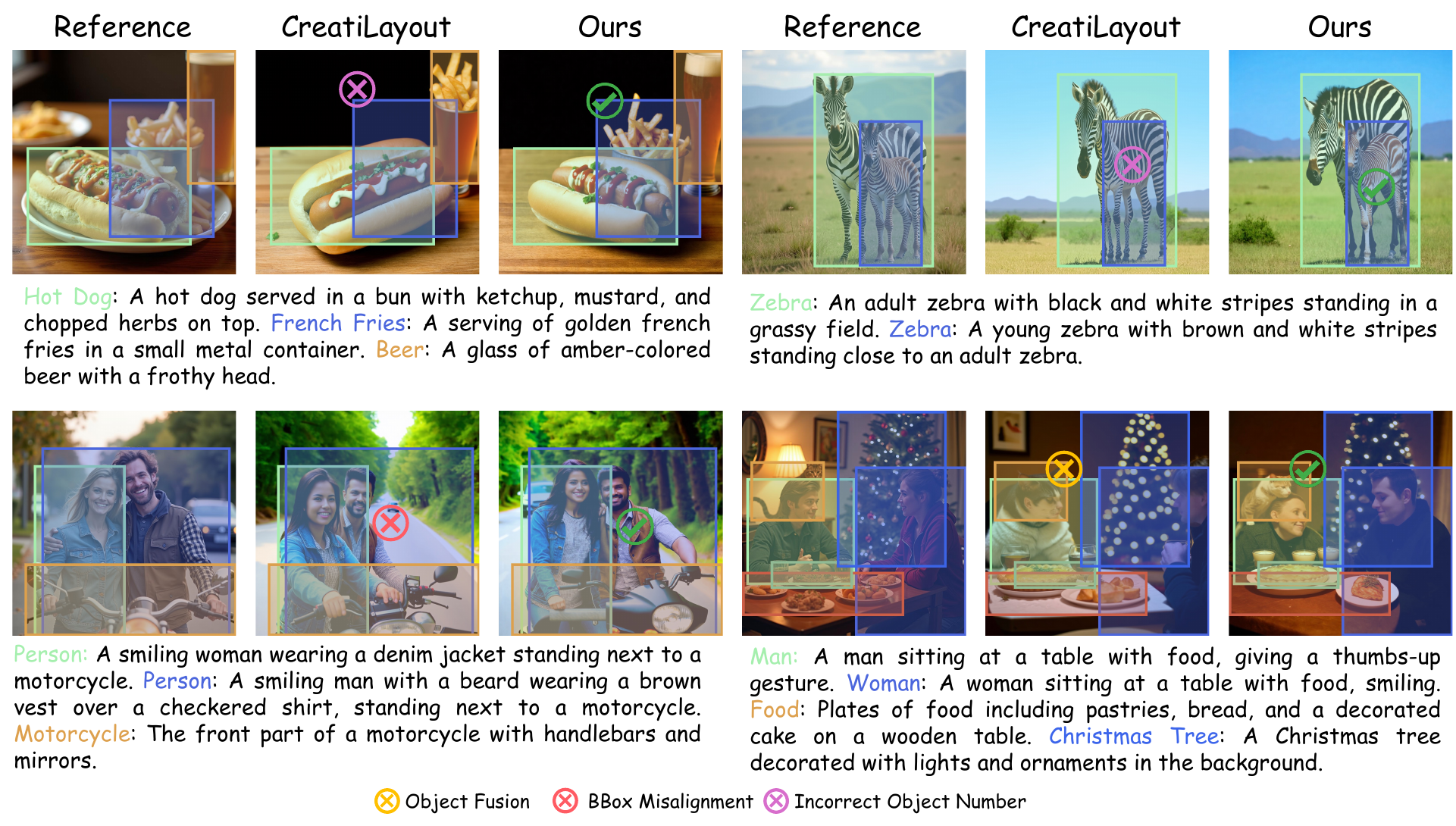}
    \caption{Comparison of generated images from CreatiLayout and CreatiLayout-AM. The CreatiLayout-AM handles overlapping instances more effectively, producing more coherent and realistic images.}
    \label{fig:CreatiLayout-AM}
\end{figure}

\section{\ourbaseline{}: Amodal Masks Improve Generation of Occluded Objects}

To address the challenges posed by complex overlaps in state-of-the-art L2I models, we incorporate amodal mask supervision during training, which provide complete object shape information even under occlusion. We construct a custom L2I training set annotated with amodal masks and train a new model to validate the effectiveness of this approach.

\paragraph{Training Dataset} We begin by synthesizing occlusions on top of FLUX-generated images. For each image, we employ Segment Anything Model v2~\citep{ravi2024sam2segmentimages} to extract amodal object masks. These masks are then used to crop individual objects, forming a pool of object-mask pairs denoted as $\mathcal{O}$. To simulate occlusions, we randomly select an object from $\mathcal{O}$ and paste it onto a target image at a location that creates overlap with an existing object. This method enables controlled occlusion synthesis while preserving the original scene context.

For each synthesized image, we use Qwen-2.5-VL-32B to generate both a global image caption and local instance descriptions, covering both original and newly pasted overlapping objects. The final training set contains approximately 67.8k images.

\paragraph{CreatiLayout-AM} Building on our curated training dataset, we introduce CreatiLayout-AM, a modified version of CreatiLayout-SD3~\citep{creatilayout} designed to enhance generation quality in the presence of occluded bounding boxes by incorporating amodal masks during training. Inspired by TokenCompose~\citep{tokencompose}, we fine-tune CreatiLayout by introducing two additional loss terms that explicitly encourage alignment between the model’s attention maps and ground truth amodal masks.

Specifically, we compute two auxiliary losses, $\mathcal{L}_{\text{token}}$ and $\mathcal{L}_{\text{pixel}}$, in addition to the original training objective. Let $\mathcal{A}^i$ denote the attention map between image tokens and the layout token corresponding to the $i^{\text{th}}$ instance, and let $m^i$ be the ground-truth amodal mask for that instance. The token-level alignment loss is defined as:
\begin{equation}
    \mathcal{L}_{\text{token}} = \frac{1}{n} \sum_{i=1}^n \left( 1 - \frac{\sum_{u} \mathcal{A}^i_{u} \cdot m^i_u}{\sum_{u} \mathcal{A}^i_{u}} \right)\label{eq:token_loss}
\end{equation}
where $n$ is the total number of instances and $u$ indexes each pixel coordinate. The pixel-level alignment loss is defined using a cross-entropy function:
\begin{equation}
    \mathcal{L}_{\text{pixel}} = \sum_{u} \text{CE}(\mathcal{A}^i_{u}, m^i_u)\label{eq:pixel_loss}
\end{equation}
where $\text{CE}$ is the cross entropy loss. The final training objective is given by:

\begin{equation}
    \mathcal{L} = \mathcal{L}_{\text{LDM}} + \lambda \mathcal{L}_{\text{token}} + \beta \mathcal{L}_{\text{pixel}} ~\label{eq:loss}
\end{equation}
where $\mathcal{L}_{\text{LDM}}$ is the original denoising loss used in the Latent Diffusion Models (LDM). 

Beyond CreatiLayout~\citep{creatilayout}, we also implemented an EliGen~\citep{eligen} based AM method, please find the detail in ~\Cref{supp:eligen_am}.
\input{tables/training_quantative_results}

\input{tables/creatilayout_vs_ours}

\section{Evaluation}
\subsection{Implementation Details}

For data curation, we use Flux-1-dev for image generation with 28 sampling steps. Qwen2.5-VL-7B is employed for image captioning, while Qwen2.5-VL-32B is used to extract bounding bboxes, instance-level captions, and relationship captions. Additionally, we apply RealVisXL\_V5.0\_Lightning\footnote{\url{https://huggingface.co/SG161222/RealVisXL_V5.0_Lightning}} for object removal during training data construction.

For CreatiLayout-AM training, we fine-tune the model for 3,500 steps on 8 NVIDIA RTX A6000 (48GB) GPUs with a batch size of 16, and a learning rate of $10^{-5}$. We use the AdamW~\citep{adamW} optimizer with bf16 precision. Please refer to the \Cref{supp:imp_details} for more details.

\subsection{Quantitative Results}

\Cref{tab:quantitative_results_all_training} presents the quantitative evaluation of multiple layout-to-image (L2I) generation methods~\citep{gligen,instancediffusion,migc,hico,creatilayout,3dis,eligen, dreamrenderer} across varying difficulty levels in the \ourbench{} benchmark: Simple, Regular, and Complex.

As task difficulty increases, all models show a noticeable decline in spatial metrics (particularly O-mIoU), highlighting the inherent challenges posed by highly overlap and semantically similar layouts. Despite this, DiT-based models demonstrate more stable visual quality and stronger semantic alignment, underscoring their robustness and scalability in handling complex generation scenarios.

\subsection{CreatiLayout-AM Comparison}

\Cref{tab:quantitative_results_creatilayout_vs_ours} demonstrates that CreatiLayout-AM outperforms the original CreatiLayout on the \textbf{Simple} and \textbf{Regular} splits, with particularly notable gains in O-mIoU ($+15.90\%$ and $+5.42\%$, respectively). These improvements are consistent across other spatial and relational metrics, including mIoU, SR\textsubscript{E}, and SR\textsubscript{R}, indicating enhanced spatial and relational alignment despite minor drops in CLIP scores. On the \textbf{Complex} split, where a distribution shift from the training set is more pronounced, performance remains competitive, exhibiting only slight declines in mIoU and CLIP. Overall, these results validate the effectiveness of amodal mask supervision in improving L2I generation under bbox overlap, presenting a promising direction for future explorations.

\subsection{Qualitative Analysis}

\begin{figure}[th]
  \centering
  \includegraphics[width=\textwidth]{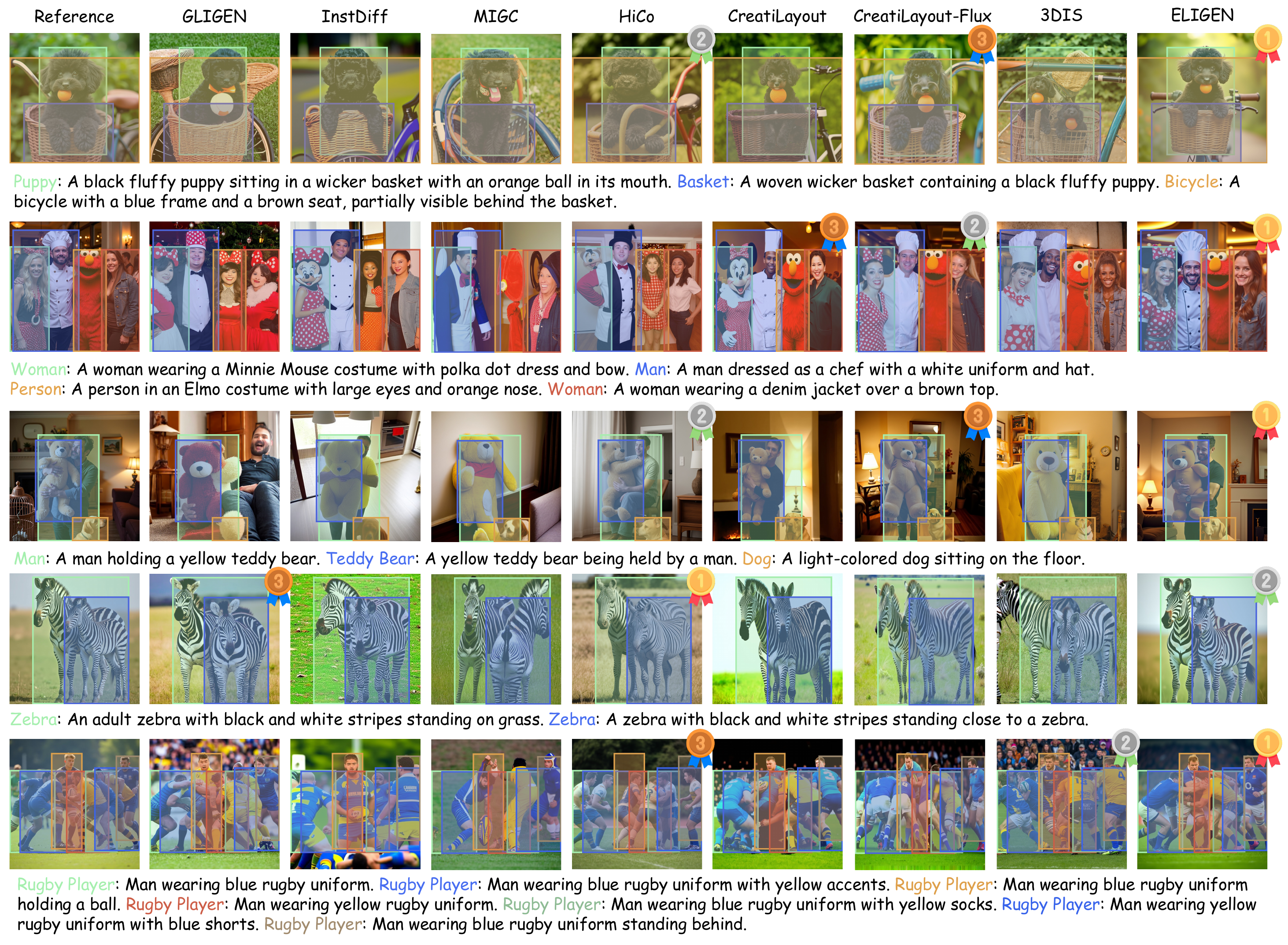}
  \caption{Comparison of generated images from different models on our \ourbench{}.}
  \label{fig:qualitative-results}
\end{figure}

For qualitative analysis, we provide a comprehensive visualization of state-of-the-art L2I models on \ourbench{}, showcasing a diverse set of examples with varying levels of layout overlap complexity in~\Cref{fig:qualitative-results}. We annotate each row with gold, silver, and bronze icons representing the top-3 performing models.

\section{Conclusion and Future Works}
In this work, we present a comprehensive study on the often-overlooked challenge of object occlusion in Layout-to-Image (L2I) generation. We introduce \textbf{\ourmetric{}}, a principal difficulty metric that captures both spatial overlap and semantic similarity, and show that higher \ourmetric{} values strongly correlate with degraded generation quality. To support rigorous evaluation, we propose \textbf{\ourbench{}}, a balanced benchmark spanning the full spectrum of layout difficulty. It features high‑fidelity images and dense captions, enabling in-depth assessment of instance interactions in densely overlapping scenes. Additionally, we demonstrate that amodal mask supervision mitigates collusion artifacts, enhancing generation quality in complex layouts. Our baseline model, \ourbaseline{}, outperforms existing methods under \ourmetric{}.

Together, our metric, benchmark, and baseline establish a unified testbed for advancing occlusion-aware, controllable image generation, and aim to inspire future methods with stronger spatial reasoning and compositional understanding.

\noindent \textbf{Acknowledgment.} This work is supported by NSF award IIS-2127544 and NSF award IIS-2433768. We thank Lambda, Inc. for their compute resource help on this project.

\clearpage

\bibliographystyle{plainnat}
\bibliography{ref}

\include{supp_arxiv}
\end{document}

%% file: tables/training_quantative_results.tex
\begin{table*}[thbp]
  \centering
  \caption{Comprehensive comparisons between training-based methods on \ourbench{}, including the newly-released model. \textbf{Bold} and \underline{underline} denote the best and the second best methods. Methods above the dashed line are U-Net–based, while those below are DiT–based. $\dagger$ means the model takes additional depth map as input.}
  \label{tab:quantitative_results_all_training}
  \setlength{\tabcolsep}{5pt}
  \setlength\dashlinedash{0.5pt}
  \setlength\dashlinegap{2pt}
  \setlength\arrayrulewidth{0.3pt}
  
  \renewcommand{\arraystretch}{1.1}
  \resizebox{\textwidth}{!}{
\begin{tabular}{@{}lccccccc@{}}
\toprule
\textbf{Method} & mIoU(\%) $\uparrow$ & O-mIoU(\%) $\uparrow$ & SR$_\text{E}$(\%) $\uparrow$ & SR$_\text{R}$(\%) $\uparrow$ & CLIP$_\text{Global}$ $\uparrow$ & CLIP$_\text{Local}$ $\uparrow$ & FID $\downarrow$ \\
\midrule

\rowcolor[RGB]{245,245,245} \multicolumn{8}{c}{\textbf{\ourbench{}-Simple}} \\
GLIGEN & 60.54$\pm$1.82 & 36.22$\pm$0.13 & 49.99$\pm$0.43 & 78.72$\pm$0.50 & 34.17$\pm$0.02 & 24.75$\pm$0.02 & 31.27$\pm$0.38 \\
InstanceDiff & \textbf{71.21}$\pm$0.11 & \textbf{49.99}$\pm$0.06 & 77.71$\pm$0.18 & 87.49$\pm$0.23 & 34.25$\pm$0.05 & 27.69$\pm$0.02 & 36.17$\pm$0.23 \\
MIGC & 58.64$\pm$0.18 & 32.15$\pm$0.31 & 63.41$\pm$0.53 & 81.60$\pm$0.19 & 33.07$\pm$0.05 & 26.49$\pm$0.10 & 31.64$\pm$0.11 \\
HiCo & 69.47$\pm$0.26 & 47.23$\pm$0.44 & 67.75$\pm$0.18 & 86.08$\pm$0.76 & 35.25$\pm$0.10 & 27.04$\pm$0.09 & 29.21$\pm$0.16 \\
\hdashline
3DIS & 65.75$\pm$0.07 & 38.38$\pm$0.23 & 86.24$\pm$0.17 & 86.98$\pm$0.24 & 35.85$\pm$0.05 & \underline{29.67}$\pm$0.03 & 29.18$\pm$0.29 \\
CreatiLayout-\tiny{SD3} & 58.78$\pm$0.44 & 32.52$\pm$0.61 & 72.34$\pm$0.61 & 84.45$\pm$0.04 & 37.29$\pm$0.04 & 27.49$\pm$0.03 & 27.51$\pm$0.15 \\
CreatiLayout-\tiny{FLUX} & \underline{71.17}$\pm$0.23 &	\underline{49.80}$\pm$0.52 &	84.35$\pm$0.49 &	\textbf{90.87}$\pm$0.27 &	\textbf{37.40}$\pm$0.15&	20.18$\pm$0.11	&\textbf{23.79}$\pm$0.17 \\
EliGen & 68.17$\pm$0.41 & 43.72$\pm$0.76 & \underline{86.50}$\pm$0.59 & 89.67$\pm$0.26 & 36.65$\pm$0.06 & 28.29$\pm$0.13 & 28.87$\pm$0.88 \\
DreamRender$^\dagger$ & 67.60$\pm$0.36 & 43.45$\pm$0.78 & \textbf{88.80}$\pm$0.44 & \underline{90.07}$\pm$0.12 & \underline{37.29}$\pm$0.03 & \textbf{30.11}$\pm$0.18 & \underline{24.91}$\pm$0.65 \\
\midrule

\rowcolor[RGB]{245,245,245} \multicolumn{8}{c}{\textbf{\ourbench{}-Regular}} \\

GLIGEN & 52.46$\pm$0.29 & 26.53$\pm$0.06 & 44.88$\pm$0.31 & 77.46$\pm$0.49 & 33.93$\pm$0.07 & 23.42$\pm$0.02 & 52.22$\pm$0.43 \\
InstanceDiff & \textbf{60.08}$\pm$0.24 & \underline{34.15}$\pm$0.16 & 72.51$\pm$0.29 & 83.36$\pm$0.30 & 33.09$\pm$0.10 & 26.19$\pm$0.03 & 59.73$\pm$0.95 \\
MIGC & 47.42$\pm$0.08 & 20.06$\pm$0.18 & 56.67$\pm$0.75 & 77.85$\pm$0.59 & 32.72$\pm$0.11 & 24.99$\pm$0.02 & 54.24$\pm$0.52 \\
HiCo & 55.02$\pm$0.33 & 29.60$\pm$0.53 & 58.24$\pm$1.01 & 79.89$\pm$0.66 & 33.91$\pm$0.14 & 25.34$\pm$0.04 & 49.07$\pm$0.45 \\
\hdashline
3DIS & 55.66$\pm$0.22 & 27.29$\pm$0.18 & \underline{80.80}$\pm$0.47 & 83.69$\pm$0.09 & 35.42$\pm$0.05 & \underline{28.12}$\pm$0.01 & 48.56$\pm$0.48 \\
CreatiLayout-\tiny{SD3} & 47.04$\pm$0.13 & 20.67$\pm$0.28 & 62.60$\pm$0.77 & 78.31$\pm$0.38 & 36.67$\pm$0.05 & 25.55$\pm$0.08 & 45.57$\pm$0.20 \\
CreatiLayout-\tiny{FLUX} & \underline{59.72}$\pm$0.29 &	\textbf{35.51}$\pm$0.44&	77.20$\pm$0.57&	\textbf{86.39}$\pm$0.31&	\underline{36.73}$\pm$0.08&	26.21$\pm$0.07&	\textbf{41.51}$\pm$0.39\\
EliGen & 58.56$\pm$0.38 & 32.62$\pm$0.52 & 80.85$\pm$0.20 & 84.42$\pm$0.36 & 36.27$\pm$0.09 & 27.05$\pm$0.12 & 45.65$\pm$1.09 \\
DreamRender$^\dagger$ & 58.08$\pm$0.36 & 33.00$\pm$0.50 & \textbf{83.52}$\pm$0.18 & \underline{84.95}$\pm$0.41 & \textbf{36.85}$\pm$0.11 & \textbf{28.74}$\pm$0.12 & \underline{42.66}$\pm$0.56 \\
\midrule
\rowcolor[RGB]{245,245,245} \multicolumn{8}{c}{\textbf{\ourbench{}-Complex}} \\
GLIGEN & 50.79$\pm$0.75 & 23.85$\pm$0.52  & 41.70$\pm$0.91 & 79.93$\pm$0.58 & 33.92$\pm$0.06 & 22.75$\pm$0.06 & 57.32$\pm$0.11 \\
InstanceDiff & \underline{53.68}$\pm$0.56 &25.63$\pm$0.34 & 66.02$\pm$0.47 & 80.34$\pm$0.25 & 32.33$\pm$0.05 & 25.53$\pm$0.01 & 66.32$\pm$0.29 \\
MIGC & 40.04$\pm$0.31 & 13.26$\pm$0.05 & 47.80$\pm$0.67 & 74.48$\pm$0.99 & 31.93$\pm$0.05 & 24.20$\pm$0.04 & 66.52$\pm$0.33 \\
HiCo & 46.56$\pm$0.31 & 20.35$\pm$0.38 & 48.88$\pm$0.32 & 75.19$\pm$0.48 & 33.15$\pm$0.18 & 24.41$\pm$0.05 & 55.78$\pm$0.35 \\
\hdashline
3DIS & 50.65$\pm$0.61 & 21.75$\pm$0.31 & 74.31$\pm$1.24 & 81.57$\pm$0.89 & 35.11$\pm$0.09 & \underline{27.35}$\pm$0.07 & 54.90$\pm$0.29 \\
CreatiLayout-\tiny{SD3} & 44.24$\pm$0.55 & 18.05$\pm$0.39 & 52.10$\pm$0.53 & 79.98$\pm$0.30 & 36.55$\pm$0.08 & 24.76$\pm$0.03 & 53.29$\pm$0.80 \\
CreatiLayout-\tiny{FLUX} & \textbf{54.50}$\pm$0.50 &	\textbf{28.97}$\pm$0.54&	69.72$\pm$0.39&	\textbf{86.45}$\pm$0.45&	\underline{36.72}$\pm$0.07	& 24.85$\pm$0.09 &	\textbf{45.66}$\pm$0.75\\
EliGen & 52.53$\pm$0.17 & \underline{26.19}$\pm$0.27 & \underline{74.03}$\pm$0.66 & 84.09$\pm$0.58 & 36.18$\pm$0.11 & 25.92$\pm$0.13 & 50.41$\pm$0.74 \\
DreamRender$^\dagger$ & 52.47$\pm$0.14 & 26.13$\pm$0.36 & \textbf{77.87}$\pm$0.45 & \underline{84.93}$\pm$0.55 & \textbf{36.75}$\pm$0.10 & \textbf{27.54}$\pm$0.11 & \underline{48.11}$\pm$0.89 \\
\bottomrule
\end{tabular}
}
\end{table*}

%% file: tables/creatilayout_vs_ours.tex
\begin{table}[htbp]
  \centering
  \caption{BaseModel v.s. Ours AM method on \ourbench{}.}
  \label{tab:quantitative_results_creatilayout_vs_ours}
  \setlength{\tabcolsep}{5pt}  
  \renewcommand{\arraystretch}{1.1}
  \setlength\dashlinedash{0.5pt}
  \setlength\dashlinegap{2pt}
\resizebox{\columnwidth}{!}{
\begin{tabular}{@{}lcccccccc@{}}
\toprule
Method       & Split & mIoU(\%) $\uparrow$ & O-mIoU(\%) $\uparrow$ & $\text{SR}_{\text{E}}$(\%) $\uparrow$ & $\text{SR}_{\text{R}}$(\%) $\uparrow$ & $\text{CLIP}_{\text{Global}}$ $\uparrow$ & $\text{CLIP}_{\text{Local}}$ $\uparrow$ & FID $\downarrow$\\ \midrule
CreatiLayout  &                          & 58.78                        & 32.52                        & 72.34                        & 84.45                        & 37.29                        & 27.49                        & 27.51                        \\
CreatiLayout-\tiny{AM}          &                          & 61.16                        & 37.69                        & 73.33                        & 84.84                        & 37.17                        & 27.44                        & 27.76                        \\
vs. BaseModel & \multirow{-3}{*}{Simple} & {\color[HTML]{009901} $+$4.05\%} & {\color[HTML]{009901} $+$15.90\%} & {\color[HTML]{009901} $+$1.37\%} & {\color[HTML]{009901} $+$0.46\%} & {\color[HTML]{FD6864} $-$0.32\%} & {\color[HTML]{FD6864} $-$0.18\%} & {\color[HTML]{FD6864} $+$0.91\%} \\
\hdashline
EliGen & & 68.17 & 43.72 & 86.50 & 89.67 & 36.65 & 28.29 & 28.87 \\
EliGen-\tiny{AM} & & 69.70 & 46.43 & 86.83 & 90.07 & 36.84 & 28.58 & 26.43 \\
vs. BaseModel & \multirow{-3}{*}{Simple} & {\color[HTML]{009901} +2.24\%} & {\color[HTML]{009901} +6.20\%} & {\color[HTML]{009901} +0.38\%} & {\color[HTML]{009901} +0.45\%} & {\color[HTML]{009901} +0.52\%} & {\color[HTML]{009901} +1.03\%} & {\color[HTML]{009901} -8.45\%} \\

\midrule

CreatiLayout  &                          & 47.04                        & 20.67                        & 62.60                        & 78.31                        & 36.67                        & 25.55                        & 45.57                        \\
CreatiLayout-\tiny{AM}          &                          & 47.38                        & 21.79                        & 63.13                        & 78.71                        & 36.49                        & 25.46                        & 46.34                        \\
vs. BaseModel & \multirow{-3}{*}{Regular} & {\color[HTML]{009901} $+$0.72\%} & {\color[HTML]{009901} $+$5.42\%} & {\color[HTML]{009901} $+$0.85\%} & {\color[HTML]{009901} $+$0.51\%} & {\color[HTML]{FD6864} $-$0.49\%} & {\color[HTML]{FD6864} $-$0.35\%} & {\color[HTML]{FD6864} $+$1.68\%} \\ 
\hdashline
EliGen & & 58.56 & 32.62 & 80.85 & 84.42 & 36.27 & 27.05 & 45.65 \\
EliGen-\tiny{AM} & & 59.44 & 33.85 & 81.18 & 85.47 & 36.46 & 27.37 & 43.52 \\
vs. BaseModel & \multirow{-3}{*}{Regular} & {\color[HTML]{009901} +1.50\%} & {\color[HTML]{009901} +3.74\%} & {\color[HTML]{009901} +0.41\%} & {\color[HTML]{009901} +2.43\%} & {\color[HTML]{009901} +0.52\%} & {\color[HTML]{009901} +1.18\%} & {\color[HTML]{009901} -4.67\%} \\
\midrule

CreatiLayout  &                          & 44.24                        & 18.05                        & 52.10                        & 79.98                        & 36.55                        & 24.76                        & 53.29                        \\
Ours          &                          & 43.97                        & 18.07                        & 52.49                        & 79.77                        & 36.32                        & 24.72                        & 53.48                        \\
vs. BaseModel & \multirow{-3}{*}{Complex}   & {\color[HTML]{FD6864} $-$0.61\%} & {\color[HTML]{009901} $+$0.11\%} & {\color[HTML]{009901} $+$0.75\%} & {\color[HTML]{FD6864} $-$0.26\%} & {\color[HTML]{FD6864} $-$0.63\%} & {\color[HTML]{FD6864} $-$0.16\%} & {\color[HTML]{FD6864} $+$0.36\%} \\
\hdashline
EliGen & & 52.53 & 26.19 & 74.03 & 84.09 & 36.18 & 25.92 & 50.41  \\
EliGen-\tiny{AM} & & 53.28 & 26.69 & 76.32 & 84.31 & 36.39 & 26.15 & 49.42  \\
vs. BaseModel & \multirow{-3}{*}{Complex} & {\color[HTML]{009901} +1.43\%} & {\color[HTML]{009901} +1.91\%} & {\color[HTML]{009901} +3.09\%} & {\color[HTML]{009901} +0.27\%} & {\color[HTML]{009901} +0.58\%} & {\color[HTML]{009901} +0.89\%} & {\color[HTML]{009901} -2.00\%} \\ \bottomrule
\end{tabular}
}
\end{table}

%% file: supp_arxiv.tex
\newpage
\begin{center}
    {\Large \textbf{Appendix}}
\end{center}
\vspace{1.0em}
\appendix

\section{Implementation Details}\label{supp:imp_details}
We begin by outlining the implementation details for the following stages: (1) data curation, (2) training, (3) inference and evaluation.

\subsection{Data Curation}

\subsubsection{Prompts Construction}

We provide the prompt templates used with Qwen-2.5-VL-7B and Qwen-2.5-VL-32B during the dataset curation process. Qwen-2.5-VL is employed for three distinct tasks: (1) Image Captioning, (2) Grounding, and (3) Relationship Extraction. The specific prompts for each task are shown below.

\lstset{basicstyle=\scriptsize\ttfamily,breaklines=true,breakautoindent=false,breakindent=0pt}

\noindent\textbf{Image Captioning}

\begin{lstlisting}[frame=single]
Give a detailed caption of this image.
\end{lstlisting}

\noindent\textbf{Grounding}

\begin{lstlisting}[frame=single]
You are required to detect the main foreground instances in the image and describe them. Response in json format:

{
    instance_category_1: {
        bbox: [x1, y1, x2, y2],
        local_prompts: description of this instance
    },
    instance_category_2: {
        bbox: [x1, y1, x2, y2],
        local_prompts: description of this instance
    },
    ...
}

Each bounding box must correspond to a single, distinct individual object - never a group
or collection. Do not merge multiple instances into one. Strictly follow this instruction
without exceptions or interpretation. Strictly follow the format in English, without any irrelevant words.
\end{lstlisting}

\noindent\textbf{Relationship Extraction}

\begin{lstlisting}[frame=single]
You are required to extract the relation between two instances in a given annotation, based on their bounding boxes and local descriptions. Only describe the relationship for the provided valid instance pairs. 
Here is the image caption: {Caption}.
Here are the instance annotations: {Bbox and Instance-Caption}.
Here are the list of valid instance pairs to describe: {Valid Overlapping Pairs}.
Response in the following json format:

{
  (Instance_1, Instance_2): Instance_1 {relationship} Instance_2,
  ...
}

where {relationship} are words in the caption and local prompts that explicitly describe the interaction or spatial relation. Each key must be a tuple from the list of valid instance pairs. If the relationship is not explicitly specified, please respond with `None'. Never invent any relationship that is not specified in the annotation. Strictly respond the relationship description in one short sentence, without any irrelevant words.
\end{lstlisting}

\subsubsection{Web-UI Construction}
\begin{figure}[h]
  \centering
  \includegraphics[width=.9\linewidth]{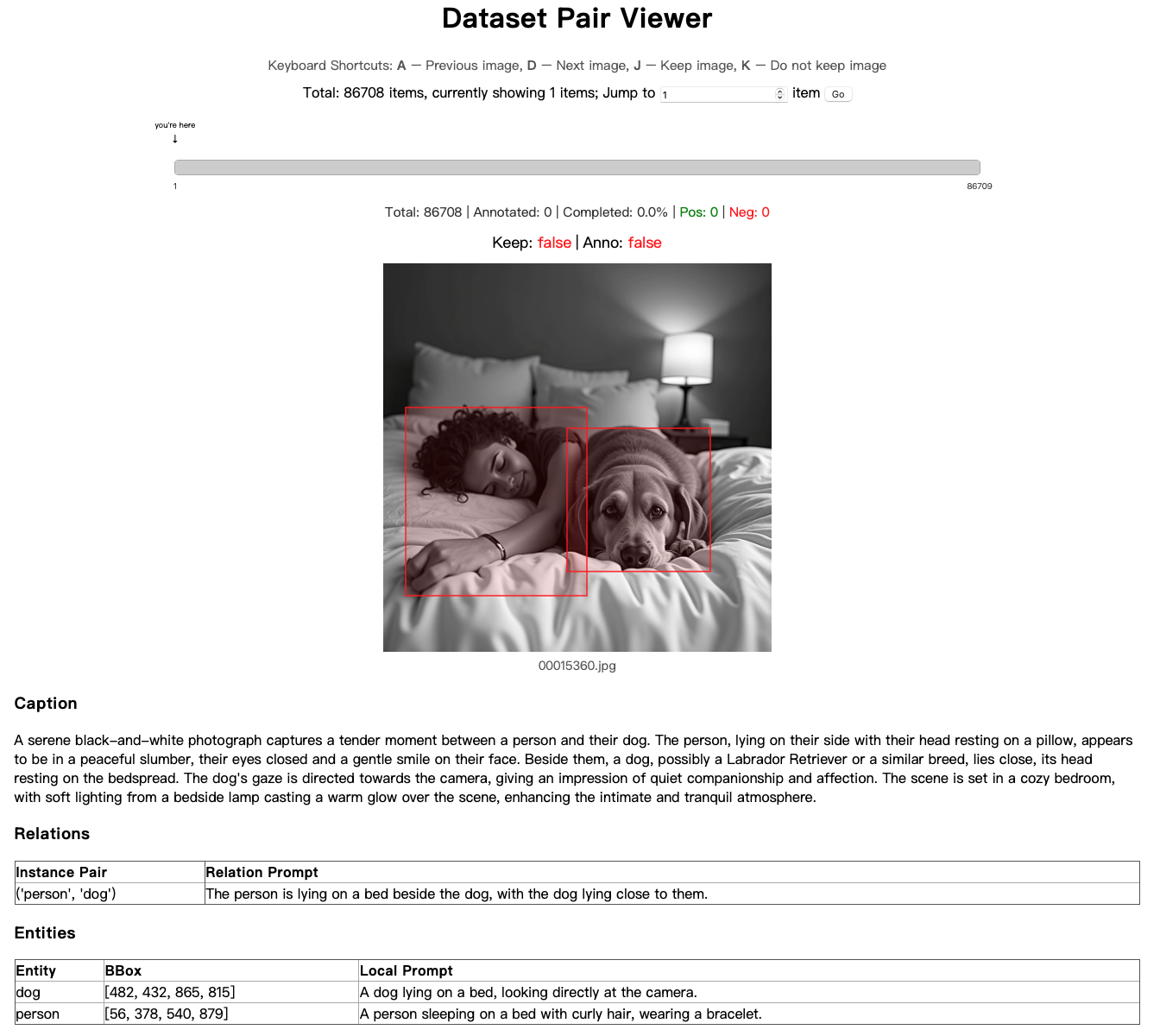}
  \caption{Web-UI used for human auditing.}
  \label{fig:webui}
\end{figure}

As part of our data curation pipeline, we incorporate human auditing to ensure data quality. To facilitate this process, we develop a custom Web-UI that displays key annotations, including image captions, bounding boxes, instance captions, and relationship captions. Annotators are instructed to verify the accuracy of each annotation to uphold high data quality standards. A screenshot of the Web-UI is provided in~\Cref{fig:webui}.

\subsection{Training}
The complete set of hyper-parameters used during training is listed in~\Cref{tab:hyperparameters}. 

\begin{table}[thbp]
  \centering
  \caption{Training hyperparameters of CreatiLayout-AM.}
  \label{tab:hyperparameters}
  \begin{tabular}{@{\hspace{3pt}}lcc@{\hspace{3pt}}}
    \toprule
    Hyperparameter           & CreatiLayout-\tiny{AM} & EliGen-\tiny{AM} \\ \midrule
    Number of GPUs           & 8 Nvidia A6000 48GB     & 8 Nvidia H100 80GB\\
    Batch Size (per GPU)     & 1     & 1\\
    Gradient Accumulation Steps & 2  & 1\\
    Gradient Checkpointing   & False & True\\
    Learning Rate            & 1e-5  & 1e-4\\
    LR Scheduler             & Linear& Linear\\
    Warm-up Steps            & 500   & 500\\
    Training Steps           & 3500  & 2000\\
    LoRA Rank                & 32    & 64\\
    Parallel                 & DDP   & FSDP \\
    $\lambda$                & 0.5   & 1\\
    $\beta$                  & 1     & 1\\ \bottomrule
  \end{tabular}
\end{table}

\subsubsection{EliGen-AM}~\label{supp:eligen_am}

To further validate the effectiveness of our training set and the proposed AM method, we conduct experiments on an additional baseline, EliGen~\citep{eligen}. Unlike CreatiLayout, which represents each entity with explicit layout tokens—thereby allowing direct extraction of entity-specific attention maps for applying our losses in \Cref{eq:loss}, EliGen constrains image tokens within a bounding box to only attend to the global description tokens and their corresponding local description tokens. This design makes it non-trivial to define the instance-level attention map required in \Cref{eq:loss}. To address this, we approximate $\mathcal{A}^i$ as the average attention map across all text tokens in the $i^{th}$ local description.

Formally, we define the attention map of $i^{th}$ instance as:
\begin{equation}
    \mathcal{\bar{A}}^i = \frac{1}{L^i}\sum_{j=0}^{L^i}\mathcal{A}^{j}
\end{equation}
where $L^i$ is the number of tokens in the $i^{th}$ local description, $\mathcal{A}^{j}$ refers to the attention map between images tokens and the $j^{th}$ local description token.

We replace the $\mathcal{A}^i$ with $\mathcal{\bar{A}}^i$ in \Cref{eq:token_loss} and \Cref{eq:pixel_loss} and train the model with the hyperparameters in~\Cref{tab:hyperparameters}. The quantitative results is shown in \Cref{tab:quantitative_results_creatilayout_vs_ours}.

\subsection{Inference and Evaluation}

For evaluation, we generate three images per method for each layout using a fixed random seed (20251202, 20251203 and 20251204) to ensure a fair comparison.

To compute the standard \textbf{mIoU}, we match each ground truth bounding box to its corresponding predicted box using the Hungarian algorithm.

For our proposed metric, \textbf{O-mIoU}, we compute the mIoU over the cropped intersection region between two instances involved in a specified relationship. We argue that this metric more effectively captures the fidelity and realism of object rendering, particularly in densely overlapping scenarios.

The CLIP score is computed using the pretrained CLIP model "ViT-B/32"~\citep{radford2021clip}.

During the inference and evaluation stages, Qwen-2.5-VL-32B is employed for \textbf{object detection} (to generate predicted bounding boxes for mIoU computation) and for \textbf{question answering}, which is used to determine the instance-level success rate \textbf{$\text{SR}_\text{E}$} and relationship-level success rate \textbf{$\text{SR}_\text{R}$}. We provide prompts for each task below.

\noindent\textbf{Object Detection}
\begin{lstlisting}[frame=single]
You are required to detect all the instances of the following categories {Categories} in the image.
Response in json format:

{
    category_1: [[x1, y1, x2, y2], [x1, y1, x2, y2], ...],
    category_2: [[x1, y1, x2, y2], ...]
    ...
}

For each category, provide a list of bounding boxes of all its instances in the image.
Each bounding box must correspond to a single, distinct individual object - never a group
or collection.
Strictly follow this instruction without exceptions or interpretation.
Strictly follow the format in English, without any irrelevant words.
\end{lstlisting}

\noindent\textbf{$\text{SR}_\text{E}$}

\begin{lstlisting}[frame=single]
You are required to answer whether the instances in an image match the corresponding descriptions, based on their bounding boxes.
Here are the instance names, the corresponding bboxes and the instance description: 
{Bbox and Instance Captions}
Please follow these rules:
Check if the generated instance visually matches its local_prompt description.
If the instance is clearly generated and not corrupted, and its key attributes described in the local_prompt are present, answer Yes.
If the instance is missing, corrupted, or the key attributes are not present, answer No.
Response in the following format:

{
  Instance_name: Yes/No,
  ...
}

Each key must be from the dict of the instance name, the corresponding bbox and the instance description.
Each value must be Yes or No.
If the instance name is not in the image, the answer should be No.
Strictly follow the format in English, without any irrelevant words.
\end{lstlisting}

\noindent\textbf{$\text{SR}_\text{R}$}

\begin{lstlisting}[frame=single]
You are required to answer whether the relationship between two instances in an image matches the description.
Here are the instance name and the bbox: {Bbox and Instance Name}.
Here are dict of the instance pair and the ground truth relationship descriptions:
{Relationships}.
Please follow these rules:
For proximity relations like near, beside, close to, next to, if the two instances are generated well (not corrupted or fused into one) and their bounding boxes are close, you can consider the description as matched.
For directional or positional relations like behind, in front of, you must strictly check
if the spatial arrangement in the image actually matches the description, because bounding boxes alone are not enough.
Response in the following format:

{
  (Instance_1, Instance_2): Yes/No,
  ...
}

Each key must be a tuple from the dict of the instance pair and the ground truth relationship descriptions.
Each value must be Yes or No.
Yes means the action/spatial relationship between the two instances matches the description.
You shouldn't pay too much attention on how well the bounding boxes are aligned.
Strictly follow the format in English, without any irrelevant words.
\end{lstlisting}

\subsection{Baseline Clarification}
We include several recent training-based L2I methods in our benchmark evaluation, some of which provide multiple variants depending on the underlying base model. To avoid ambiguity, we clarify our choices here. For HiCo~\citep{hico}, we use the HiCo-SD1.5 model, as the SDXL version was not publicly available at the time of evaluation. For 3DIS~\citep{3dis} and DreamRender~\citep{dreamrenderer}, we adopt the FLUX-based versions. In addition, since DreamRender requires auxiliary modalities such as canny edge maps and depth maps, we utilize DepthAnything v2~\citep{depthanythingv2} to extract depth maps from reference images, which are then provided together with the layout as inputs to the model.

\section{Quantitative Results Analysis on Training-Free Methods}

In addition to the training-based approaches, we evaluate several training-free methods on \ourbench{}~\citep{boxdiff,layout-guidance,attn-refocusing,groundit,trainigfreeregional}, as shown in \Cref{tab:quantitative_results_all_supp_training_free}.

Specifically, on the \textbf{OverLayBench-Simple} split, RegionalPrompting achieves the highest overall performance, with the best mIoU (42.54\%), O-mIoU (20.10\%), SR$_\text{E}$ (73.49\%), SR$_\text{R}$ (75.81\%), CLIP$_\text{Local}$ (27.40), and FID (23.94). On the \textbf{Regular} split, it continues to lead in mIoU (32.72\%), O-mIoU (12.29\%), SR$_\text{E}$ (63.74\%), and CLIP$_\text{Local}$ (25.82), again achieving the lowest FID (43.13), despite a slight drop in SR$_\text{R}$ (67.08\%). For the more challenging \textbf{Complex} split, RegionalPrompting still attains the highest mIoU (28.35\%), O-mIoU (9.05\%), SR$_\text{E}$ (53.56\%), and CLIP$_\text{Local}$ (25.29), alongside a competitive FID (49.41), further demonstrating its robustness in dense and overlap-heavy scenarios.

\input{supp_tables/training_free_quantative_results}

\begin{figure}[htb]
  \centering
  \includegraphics[width=1\linewidth]{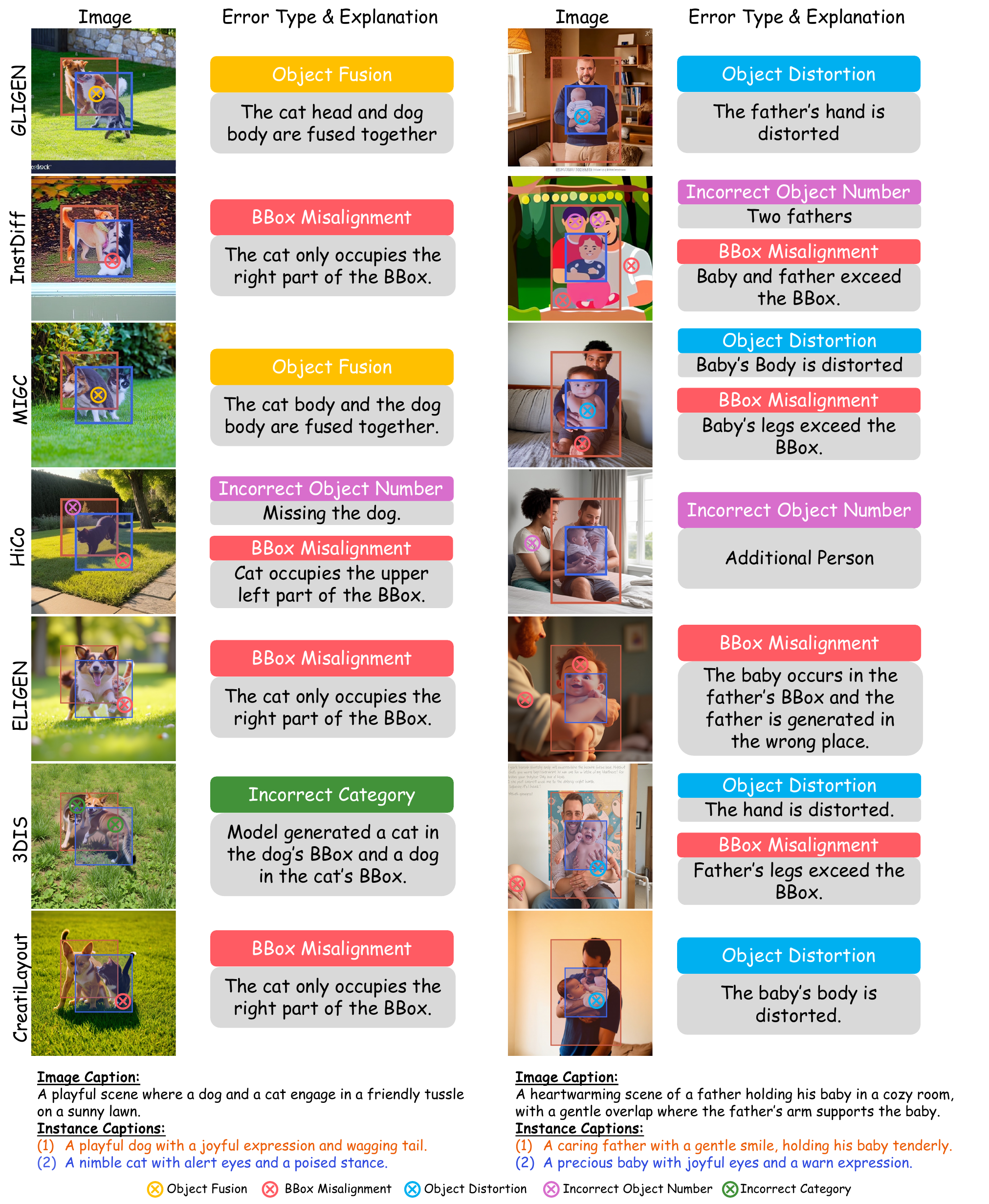}
  \caption{Error pattern analysis and explanation for existing models.}
  \label{fig:failure_case_detail}
\end{figure}

\section{Error Pattern Analysis of Existing Methods}\label{supp:failure_case}
\begin{figure}[thb]
  \centering
  \includegraphics[width=1\linewidth]{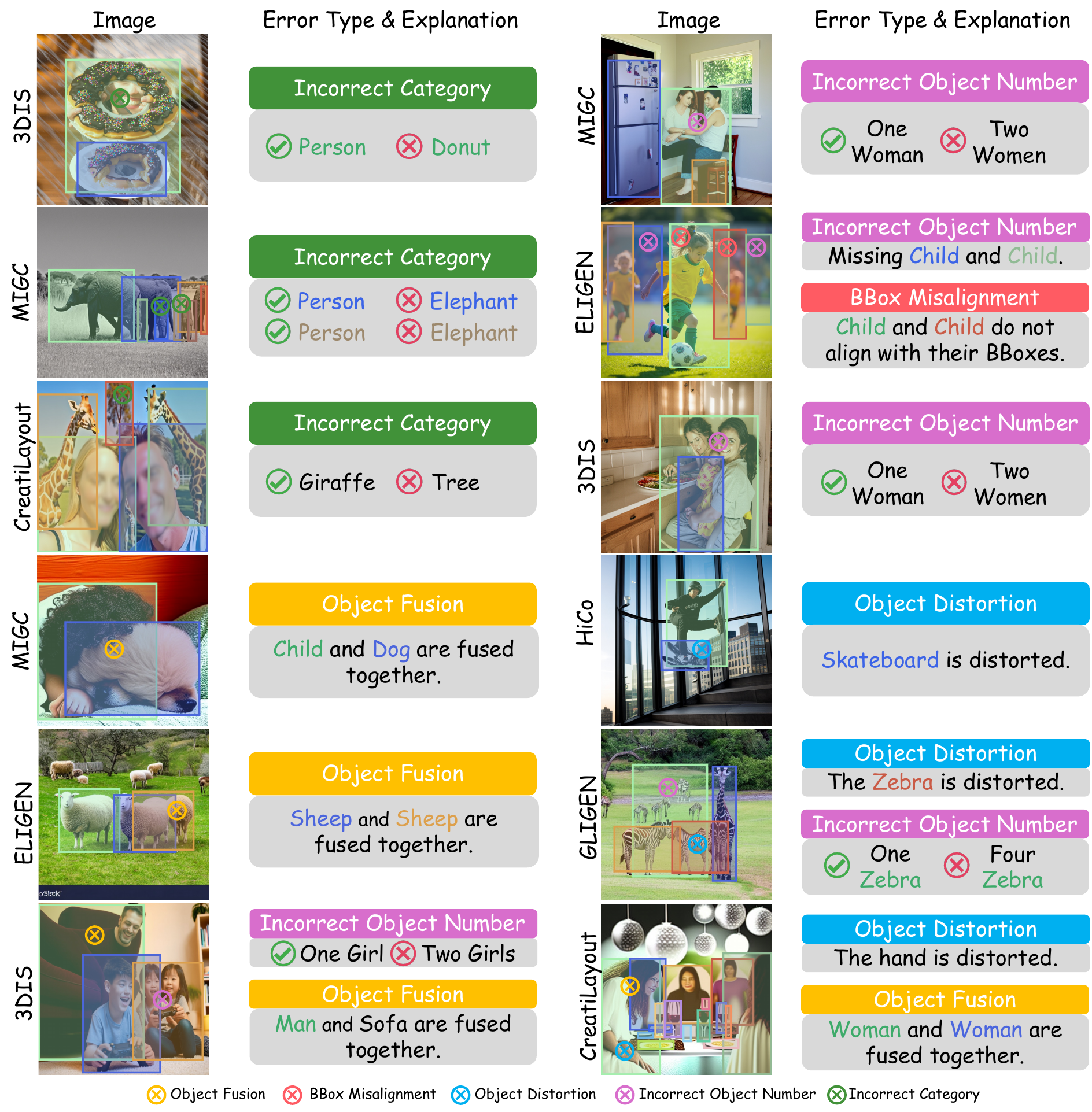}
  \caption{More examples of error patterns for existing methods.}
  \label{fig:error_pattern}
\end{figure}

We identify and categorize the common failure patterns observed in existing methods into five major classes: 1) \textbf{Incorrect Object Number}, where models either hallucinate additional undesired objects or fail to generate required instances within the specified bounding box regions; 2) \textbf{Object Fusion}, where models struggle to generate distinct instances for overlapping bounding boxes, instead producing a single merged or entangled object; 3) \textbf{Object Distortion}, where the generated instance lacks realism, often exhibiting severe deformation or artifacts that degrade perceptual quality; 4) \textbf{Incorrect Category}, where the generated object does not match the intended category, undermining semantic correctness; 5) \textbf{BBox Misalignment}, where the object does not properly align with its designated bounding box, either overflowing beyond the box or failing to fully occupy the allocated region, thus breaking spatial consistency. Please refer to~\Cref{fig:failure_case_detail,fig:error_pattern} for additional visual examples and explanations of each error pattern.

\section{Comparison Between GroundingDINO and Qwen}
\begin{figure}[thb]
  \centering
  \includegraphics[width=1\linewidth]{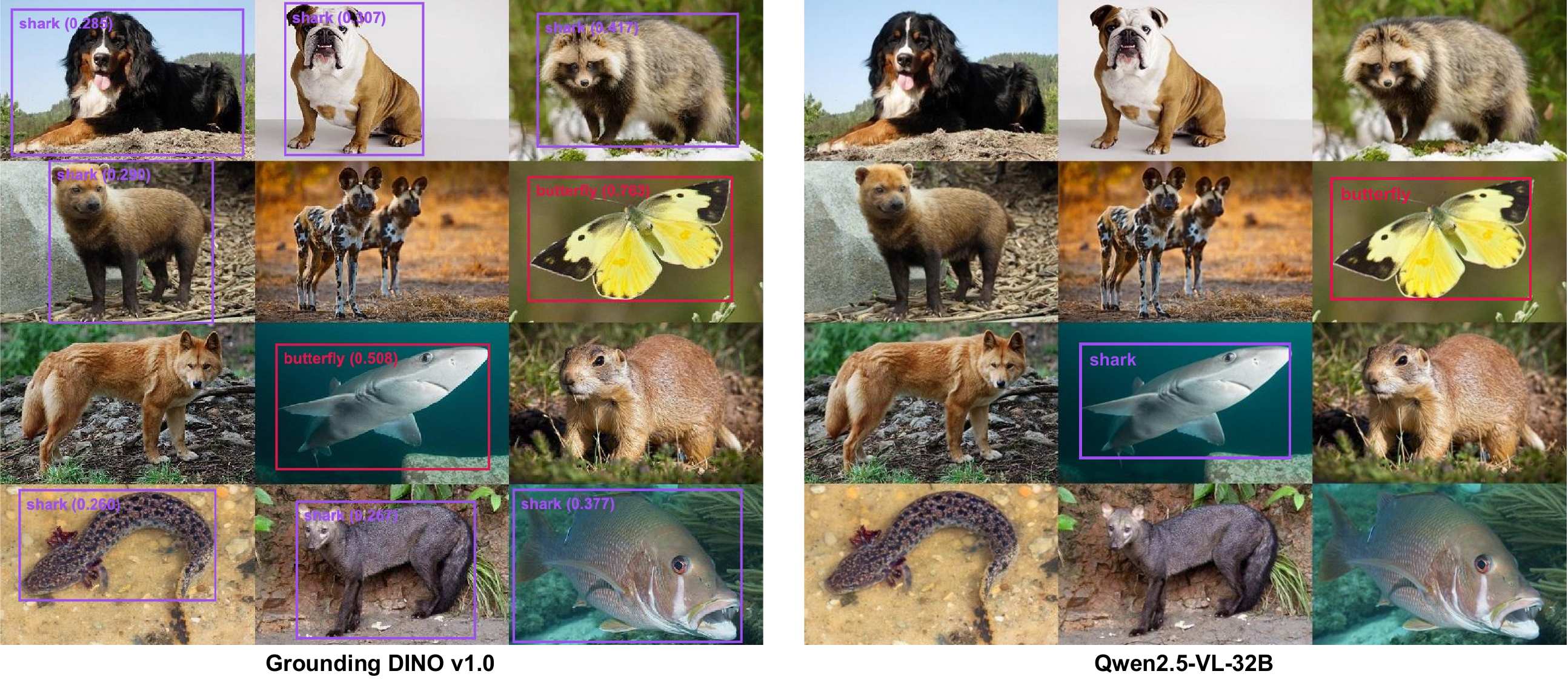}
  \caption{Comparison of grounding performance between GroundingDINO v1.0 and Qwen-2.5-VL-32B.}
  \label{fig:groundDINO_vs_QWen}
\end{figure}

We compare the grounding capabilities of GroundingDINO v1.0~\cite{liu2023groundingdino} and Qwen-2.5-VL-32B~\cite{qwen} using an example from GroundingDINO’s official demo\footnote{\url{https://cloud.deepdataspace.com/playground/grounding_dino}}. As illustrated in~\Cref{fig:groundDINO_vs_QWen}, GroundingDINO produces multiple false positives, including a misclassification of a shark as a butterfly. In contrast, Qwen demonstrates more accurate and robust detection performance.

\section{User Study}\label{supp:user_study}
We conducted a user study with 15 participants over 60 image pairs (see \Cref{tab:user-study}), comparing CreatiLayout and CreatiLayout-AM across three difficulty levels. Excluding “No Preference” cases, our method achieved winning rates of 55.2\%, 51.9\%, and 46.8\% on Simple, Regular, and Complex settings, respectively, showing moderate preference in simpler scenarios.

\begin{table*}[hb]
  \centering
  \caption{User study results across three difficulty levels.}
  \label{tab:user-study}
\begin{tabular}{lccc}
\toprule
Ours vs. CreatiLayout & Simple & Regular & Complex \\
\midrule
Winning Rate (\%) & 55.2 & 51.9 & 46.8 \\
\bottomrule
\end{tabular}
\end{table*}

%% file: supp_tables/training_free_quantative_results.tex
\begin{table*}[h]
  \centering
  \caption{Comprehensive comparison between training-free methods on \ourbench{}. \textbf{Bold} and \underline{underline} denote the best and the second best methods. Methods above the dashed line are U-Net–based, while those below are DiT–based.}
  \label{tab:quantitative_results_all_supp_training_free}
  \setlength{\tabcolsep}{5pt}
  \setlength\dashlinedash{0.5pt}
  \setlength\dashlinegap{2pt}
  \setlength\arrayrulewidth{0.3pt}
  
  \renewcommand{\arraystretch}{1.1}
  \resizebox{\textwidth}{!}{
\begin{tabular}{@{}lccccccc@{}}
\toprule
\textbf{Method} & mIoU(\%) $\uparrow$ & O-mIoU(\%) $\uparrow$ & SR$_\text{E}$(\%) $\uparrow$ & SR$_\text{R}$(\%) $\uparrow$ & CLIP$_\text{Global}$ $\uparrow$ & CLIP$_\text{Local}$ $\uparrow$ & FID $\downarrow$ \\
\midrule

\rowcolor[RGB]{245,245,245} \multicolumn{8}{c}{\textbf{\ourbench{}-Simple}} \\

BoxDiff & 24.48$\pm$0.08 & 7.71$\pm$0.43 & 42.03$\pm$0.89 & 69.94$\pm$0.52 & \textbf{36.78}$\pm$0.06 & 21.33$\pm$0.08 & \underline{34.65}$\pm$0.20 \\
LayoutGuidance & 23.12$\pm$0.27 & 7.92$\pm$0.13 & 45.78$\pm$0.65 & 70.83$\pm$0.70 & 33.47$\pm$0.10 & 21.22$\pm$0.01 & 74.40$\pm$1.01 \\
R\&B & 27.78$\pm$1.90 & 9.70$\pm$1.16 & 36.98$\pm$0.52 & 64.05$\pm$0.45 & 34.64$\pm$0.18 & 21.32$\pm$0.04 & 36.57$\pm$2.30 \\
\hdashline

GroundDiT & \underline{31.92}$\pm$0.34 & \underline{10.93}$\pm$0.36 & \underline{48.57}$\pm$0.89 & \underline{75.26}$\pm$0.40 & \underline{36.10}$\pm$0.12 & \underline{22.73}$\pm$0.03 & 34.98$\pm$0.63 \\
RegionalPrompting & \textbf{42.54}$\pm$0.17 & \textbf{20.10}$\pm$0.27 & \textbf{73.49}$\pm$0.53 & \textbf{75.81}$\pm$0.32 & 35.45$\pm$0.08 & \textbf{27.40}$\pm$0.06 & \textbf{23.94}$\pm$0.10 \\

\midrule

\rowcolor[RGB]{245,245,245} \multicolumn{8}{c}{\textbf{\ourbench{}-Regular}} \\
BoxDiff & 19.40$\pm$0.30 & 5.33$\pm$0.24 & 37.58$\pm$0.36 & 71.81$\pm$0.44 & \textbf{36.50}$\pm$0.05 & 19.97$\pm$0.03 & 55.41$\pm$0.28 \\
LayoutGuidance & 15.81$\pm$0.20 & 4.51$\pm$0.19 & \underline{44.94}$\pm$0.07 & \underline{72.76}$\pm$0.86 & 31.76$\pm$0.01 & 20.05$\pm$0.02 & 128.16$\pm$0.35 \\
R\&B & 20.35$\pm$0.74 & 5.54$\pm$0.61 & 32.85$\pm$0.42 & 65.01$\pm$0.06 & 34.49$\pm$0.14 & 19.88$\pm$0.04 & 58.55$\pm$1.79 \\
\hdashline
GroundDiT & \underline{24.03}$\pm$0.20 & \underline{6.70}$\pm$0.31 & 42.24$\pm$0.66 & \textbf{74.02}$\pm$0.46 & \underline{36.00}$\pm$0.12 & \underline{21.10}$\pm$0.05 & \underline{54.94}$\pm$0.49 \\
RegionalPrompting & \textbf{32.72}$\pm$0.30 & \textbf{12.29}$\pm$0.23 & \textbf{63.74}$\pm$0.73 & 67.08$\pm$0.69 & 34.46$\pm$0.10 & \textbf{25.82}$\pm$0.10 & \textbf{43.13}$\pm$0.31 \\
\midrule
\rowcolor[RGB]{245,245,245} \multicolumn{8}{c}{\textbf{\ourbench{}-Complex}} \\
BoxDiff & 20.02$\pm$0.32 & 5.21$\pm$0.15 & 33.52$\pm$0.85 & \underline{76.41}$\pm$0.51 & \textbf{36.92}$\pm$0.02 & 19.91$\pm$0.03 & 59.70$\pm$0.67 \\
LayoutGuidance & 16.34$\pm$0.33 & 4.01$\pm$0.14 & 37.76$\pm$0.91 & 75.53$\pm$1.41 & 32.75$\pm$0.04 & 19.72$\pm$0.05 & 119.32$\pm$1.19 \\
R\&B & 19.80$\pm$0.67 & 4.85$\pm$0.36 & 28.38$\pm$1.05 & 69.97$\pm$1.31 & 34.57$\pm$0.19 & 19.47$\pm$0.06 & 63.71$\pm$2.82 \\
\hdashline

GroundDiT & \underline{24.77}$\pm$0.88 & \underline{6.58}$\pm$0.31 & \underline{37.85}$\pm$1.27 & \textbf{77.03}$\pm$1.81 & \underline{36.20}$\pm$0.24 & \underline{20.55}$\pm$0.06 & \underline{55.59}$\pm$1.39 \\ 
RegionalPrompting & \textbf{28.35}$\pm$0.94 & \textbf{9.05}$\pm$0.59 & \textbf{53.56}$\pm$1.58 & 60.37$\pm$1.56 & 33.34$\pm$0.13 & \textbf{25.29}$\pm$0.05 & \textbf{49.41}$\pm$1.25 \\

\bottomrule
\end{tabular}
 }

\end{table*}